\documentclass[11pt]{article}

\usepackage[margin=1in]{geometry}
\usepackage{amsmath,amssymb,amsfonts}
\usepackage{graphicx}
\usepackage{booktabs}
\usepackage{multirow}
\usepackage{hyperref}
\usepackage{xcolor}
\usepackage{subcaption}
\usepackage{authblk}
\usepackage{bm}
\usepackage{float}
\usepackage{microtype}

\title{\textbf{HamVision: Hamiltonian Dynamics as Inductive Bias for\\Medical Image Analysis}}
\author[1]{Mohamed Mabrok}
\affil[1]{Department of Mathematics and Statistics, Qatar University, Doha, Qatar}
\date{}

\begin{document}
\maketitle

\begin{abstract}
We present HamVision, a framework for medical image analysis that uses the damped harmonic oscillator, a fundamental building block of signal processing, as a structured inductive bias for both segmentation and classification tasks. The oscillator's phase-space decomposition yields three functionally distinct representations: position~$q$ (feature content), momentum~$p$ (spatial gradients that encode boundary and texture information), and energy $H = \tfrac{1}{2}|z|^2$ (a parameter-free saliency map). These representations emerge from the dynamics, not from supervision, and can be exploited by different task-specific heads without any modification to the oscillator itself. For segmentation, energy gates the skip connections while momentum injects boundary information at every decoder level (HamSeg). For classification, the three representations are globally pooled and concatenated into a phase-space feature vector (HamCls). We evaluate HamVision across ten medical imaging benchmarks spanning five imaging modalities. On segmentation, HamSeg achieves state-of-the-art Dice scores on ISIC\,2018 (89.38\%), ISIC\,2017 (88.40\%), TN3K (87.05\%), and ACDC (92.40\%), outperforming most baselines with only 8.57M parameters. On classification, HamCls achieves state-of-the-art accuracy on BloodMNIST (98.85\%) and PathMNIST (96.65\%), and competitive results on the remaining MedMNIST datasets against MedMamba and MedViT. Diagnostic analysis confirms that the oscillator's momentum consistently encodes an interior$\,{>}\,$boundary$\,{>}\,$exterior gradient for segmentation and that the energy map correlates with discriminative regions for classification, properties that emerge entirely from the Hamiltonian dynamics. Code is available at \url{https://github.com/Minds-R-Lab/hamvision}.
\end{abstract}

\section{Introduction}
\label{sec:introduction}

Medical image analysis encompasses a broad range of tasks, from pixel-level segmentation of anatomical structures to image-level classification of pathological conditions. These tasks differ in their outputs, dense spatial maps versus scalar labels, but share a fundamental requirement: the ability to extract discriminative features from complex, noisy images where the relevant information may be subtle gradients at tissue boundaries, textural patterns indicative of disease, or spatial distributions of intensity. The dominant paradigm addresses each task with specialized architectures: U-Nets with skip connections for segmentation~\cite{ronneberger2015unet}, convolutional or transformer-based backbones with pooling heads for classification~\cite{he2016resnet,dosovitskiy2021vit}. Both paradigms learn generic feature transformations from data, relying on model capacity rather than structural priors to discover what computations are useful.

This paper takes a different approach. We observe that the damped harmonic oscillator is a universal signal analyzer whose intrinsic dynamics produce three representations that are useful for \emph{any} medical image analysis task. When a spatially varying signal, a row or column of an image feature map, drives an oscillator with natural frequency~$\omega$ and damping~$\nu$, the oscillator's response decomposes into position~$q$ (filtered signal content), momentum~$p = \dot{q}$ (the rate of spatial change, which peaks at boundaries and vanishes in homogeneous regions), and energy~$H = \tfrac{1}{2}(q^2 + p^2)$ (a scalar measure of local excitation that integrates both signals into a spatial saliency map). None of these properties are learned; they are intrinsic to the dynamics. A bank of oscillators at different frequencies forms a neural filterbank~\cite{lyon2017human}, decomposing features into frequency bands with per-channel momentum and energy as structured byproducts.

The key insight of this work is that these three representations, position, momentum, energy, are sufficient to build both a segmentation head and a classification head, without modifying the oscillator itself. For segmentation, boundaries are the critical information: momentum detects them (large~$|p|$ at transitions), energy gates them (highlighting salient regions), and both propagate through the decoder at all resolutions. For classification, discriminative power lies in global statistics: the distribution of momentum magnitudes captures texture complexity, energy captures overall image activity, and position captures content, all pooled into a compact classification vector. The oscillator bottleneck is a shared computational primitive; only the downstream head changes between tasks.

This constitutes a structured inductive bias in the same sense that convolutions impose translation equivariance or graph networks impose permutation equivariance. Our bias is that features should evolve according to damped oscillator dynamics, which implicitly encodes the prior that spatially varying signals contain boundary, frequency, and saliency information that is useful for medical image analysis. The analogy to Hamiltonian Monte Carlo~\cite{neal2011hmc} is instructive: that method uses oscillator dynamics not because sampling is a physical process, but because the dynamics have useful exploration properties. We use Hamiltonian dynamics not because images are physical systems, but because the resulting phase-space decomposition provides precisely the representations that both segmentation and classification require.

Mamba-3~\cite{mamba3} has independently validated that complex-valued state transitions enable capabilities that real-valued SSMs provably lack, confirming the algebraic necessity of rotational dynamics for certain computational tasks. Our work leverages this validation by showing that the \emph{physical interpretation} of the complex state, the decomposition into position, momentum, and energy, provides practical and interpretable structured representations for both segmentation and classification in vision tasks.

This paper makes four contributions. First, we introduce a Hamiltonian bottleneck that produces position, momentum, and energy as structured intermediate representations from a single dynamical process. Second, we design two task-specific heads that exploit these representations: energy-gated skip connections with momentum injection for segmentation, and phase-space pooling for classification. Third, we demonstrate that HamVision, a single architecture with a shared oscillator bottleneck, achieves state-of-the-art results across ten benchmarks spanning five imaging modalities and two distinct tasks, with only 7--9M parameters. Fourth, we provide diagnostic evidence that the oscillator's outputs carry interpretable, task-relevant information that emerges from the dynamics without explicit supervision.

\section{Related Work}
\label{sec:related}

The U-Net~\cite{ronneberger2015unet} established the encoder--decoder paradigm with skip connections for biomedical segmentation, subsequently extended by attention gates~\cite{oktay2018attention}, dense connections~\cite{huang2020unet3plus}, and vision transformers that provide global context at the cost of quadratic complexity~\cite{chen2021transunet,cao2022swinunet}. For classification, ResNets~\cite{he2016resnet} remain the standard backbone, with MedViT~\cite{medvit2023} adapting vision transformers for medical imaging and MedMamba~\cite{yue2024medmamba} introducing state-space models for efficient classification. Our framework, HamVision, preserves the U-Net topology for segmentation and a simple pooling head for classification, replacing the generic bottleneck with a physics-structured alternative.

Structured state-space models, beginning with S4~\cite{gu2022s4} and culminating in Mamba~\cite{gu2023mamba}, have emerged as efficient alternatives to attention for modeling long-range dependencies. VMamba~\cite{liu2024vmamba} extended this to two-dimensional data through multi-directional scanning. In medical imaging, VM-UNet~\cite{ruan2024vmunet}, U-Mamba~\cite{ma2024umamba}, and FreqConvMamba~\cite{freqconvmamba2026} embed generic state-space blocks within U-Net variants for segmentation, while MedMamba~\cite{yue2024medmamba} uses them for classification. These methods all treat the state transition as a learned diagonal with no structural constraints. Our work constrains the transition to follow oscillator dynamics, exploiting the rich physical structure of the resulting phase space.

The use of dynamical systems as computational primitives has a long history. Neural ODEs~\cite{chen2018neuralode} parameterize hidden state evolution as continuous dynamics. Hamiltonian Neural Networks~\cite{greydanus2019hamiltonian} learn energy-conserving systems from data. Oscillatory networks have been studied for synchronization-based grouping~\cite{wang1995oscillatory} and pattern recognition~\cite{hoppensteadt1999oscillatory}, while unitary RNNs~\cite{arjovsky2016unitary} use rotation matrices, a special case of oscillator dynamics, to combat vanishing gradients. Mamba-3~\cite{mamba3} independently proved that complex-valued transitions enable state-tracking via data-dependent rotary embeddings. Our work is the first to exploit the full phase-space decomposition as structured intermediate representations for vision, and the first to demonstrate that these representations are useful for both segmentation and classification from a single shared oscillator.

\section{Method}
\label{sec:method}

\subsection{Hamiltonian Mechanics of the Damped Oscillator}

We begin from first principles. Consider a unit-mass particle attached to a spring with stiffness $k > 0$, subject to viscous damping $\nu > 0$ and driven by an external force $u(t)$. The particle has position $q(t)$ and momentum $p(t) = \dot{q}(t)$. The Hamiltonian of the conservative system (no friction) is the total energy:
\begin{equation}
    \mathcal{H}(q, p) = \underbrace{\tfrac{1}{2}p^2}_{\text{kinetic}} + \underbrace{\tfrac{1}{2}kq^2}_{\text{potential}},
    \label{eq:hamiltonian}
\end{equation}
and Hamilton's equations of motion are $\dot{q} = \partial\mathcal{H}/\partial p = p$ and $\dot{p} = -\partial\mathcal{H}/\partial q = -kq$. A conservative oscillator preserves energy indefinitely ($\dot{\mathcal{H}} = 0$), which is undesirable for sequence modeling, the system would retain all information from all past inputs with equal weight.

To introduce a controllable forgetting mechanism, we add a non-conservative dissipative force $F_\text{damp} = -\nu p$ that acts against the direction of motion, yielding the full equations of motion:
\begin{equation}
    \begin{cases}
    \dot{q}(t) = p(t) \\
    \dot{p}(t) = -kq(t) - \nu p(t) + u(t)
    \end{cases}
    \label{eq:eom}
\end{equation}
In state-space form with $\mathbf{x} = [q, p]^\top$, this becomes $\dot{\mathbf{x}} = \mathbf{A}\mathbf{x} + \mathbf{B}u$ where
\begin{equation}
    \mathbf{A} = \begin{pmatrix} 0 & 1 \\ -k & -\nu \end{pmatrix}, \quad \mathbf{B} = \begin{pmatrix} 0 \\ 1 \end{pmatrix}.
    \label{eq:ssm_matrix}
\end{equation}
This matrix is not learned arbitrarily; it is structurally constrained to represent a damped harmonic oscillator. The energy dissipation rate follows directly:
\begin{equation}
    \frac{d\mathcal{H}}{dt} = p\dot{p} + kq\dot{q} = p(-kq - \nu p + u) + kqp = -\nu p^2 + pu.
    \label{eq:energy_rate}
\end{equation}
In the absence of driving ($u = 0$), energy dissipates at rate $\dot{\mathcal{H}} = -\nu p^2 \leq 0$, guaranteeing that the system cannot diverge. This is a structural stability property inherited from the physics, not an architectural trick.

\paragraph{Symplectic structure and phase-space geometry.}
The conservative oscillator ($\nu = 0$, $u = 0$) is a symplectic map: the flow $\phi_t: (q_0, p_0) \mapsto (q_t, p_t)$ preserves the symplectic 2-form $\mathrm{d}q \wedge \mathrm{d}p$, or equivalently, the phase-space area element. This is a consequence of Liouville's theorem: the Jacobian of the Hamiltonian flow satisfies $\det(\partial \phi_t / \partial \mathbf{x}_0) = 1$ for all $t$. Geometrically, the conservative oscillator traces out ellipses in the $(q, p)$ plane with semi-axes $\sqrt{2\mathcal{H}/k}$ and $\sqrt{2\mathcal{H}}$, and the orbit period $T = 2\pi/\omega$ determines the temporal scale of the filter.

When damping is introduced ($\nu > 0$), the symplectic structure is broken: the flow contracts phase-space volume at a rate $\nabla \cdot \dot{\mathbf{x}} = \text{tr}(\mathbf{A}) = -\nu < 0$, so that the phase-space area shrinks exponentially as $e^{-\nu t}$. This contraction is precisely the forgetting mechanism: information about past inputs is geometrically compressed in phase space at a rate controlled by $\nu$. The interplay between area-preserving rotation (from the Hamiltonian) and area-contracting dissipation (from damping) is what gives the oscillator its distinctive character as a frequency-selective, memory-decaying filter.

\paragraph{Lyapunov stability analysis.}
To establish formal stability, we use the Hamiltonian $\mathcal{H}$ as a Lyapunov function. For the undriven system ($u = 0$), $\mathcal{H}$ is positive definite ($\mathcal{H} > 0$ for $(q,p) \neq (0,0)$) and radially unbounded ($\mathcal{H} \to \infty$ as $\|(q,p)\| \to \infty$). Its time derivative satisfies $\dot{\mathcal{H}} = -\nu p^2 \leq 0$, which is negative semi-definite. By LaSalle's invariance principle, every trajectory converges to the largest invariant set contained in $\{(q,p) : p = 0\}$. On this set, $\dot{q} = p = 0$ and $\dot{p} = -kq = 0$ imply $q = 0$, so the origin is globally asymptotically stable. For the driven system, the energy satisfies $\dot{\mathcal{H}} = -\nu p^2 + pu \leq -\nu p^2 + |p||u| \leq -\frac{\nu}{2}p^2 + \frac{1}{2\nu}u^2$ (by Young's inequality), yielding the bound:
\begin{equation}
    \mathcal{H}(t) \leq e^{-\nu t}\mathcal{H}(0) + \frac{1}{2\nu^2}\int_0^t e^{-\nu(t-s)} u(s)^2 \, ds,
    \label{eq:lyapunov_bound}
\end{equation}
which proves that energy remains bounded whenever the input is bounded (BIBO stability) and decays exponentially to a level proportional to $\|u\|^2 / \nu^2$. The decay rate $\nu$ thus controls the trade-off between memory length and stability margin.

\subsection{Complex Diagonalization and Phase-Space Transformation}

The coupled system~\eqref{eq:eom} is computationally inefficient for parallel training because $q$ and $p$ are coupled. We diagonalize by computing the eigenvalues of $\mathbf{A}$ via the characteristic polynomial $\lambda^2 + \nu\lambda + k = 0$, yielding:
\begin{equation}
    \lambda_{\pm} = -\frac{\nu}{2} \pm \frac{1}{2}\sqrt{\nu^2 - 4k}.
    \label{eq:eigenvalues}
\end{equation}
Three regimes arise depending on the discriminant $\Delta = \nu^2 - 4k$. When $\Delta < 0$ (underdamped, $\nu^2 < 4k$), the eigenvalues are complex conjugates $\lambda_{\pm} = -\nu/2 \pm i\omega_d$ where $\omega_d = \sqrt{k - \nu^2/4}$ is the damped natural frequency. This is the regime of interest: the system oscillates while decaying. In the small-damping approximation ($\nu^2 \ll 4k$), we have $\omega_d \approx \omega = \sqrt{k}$ and the eigenvalues simplify to $\lambda \approx -\nu + i\omega$.

Since $\mathbf{A}$ is real-valued, its eigenvalues and eigenvectors come in conjugate pairs. We project the physical state $\mathbf{x} = [q, p]^\top$ onto the eigenvector basis, obtaining a complex scalar $z(t) = q(t) + ip(t)$ that satisfies the decoupled equation:
\begin{equation}
    \frac{dz}{dt} = (-\nu + i\omega)z + u(t).
    \label{eq:complex_ode}
\end{equation}
This transformation is the cornerstone of our approach: scalar complex multiplication is commutative and associative, enabling efficient parallelization via cumulative sums. Crucially, the real and imaginary parts of $z$ retain their physical meaning: $\text{Re}(z) = q$ is the position (filtered feature content) and $\text{Im}(z) = p$ is the momentum (spatial derivative of features). The squared modulus $|z|^2 = q^2 + p^2 = 2\mathcal{H}$ is proportional to the total energy.

\paragraph{Explicit eigenvector construction.}
The diagonalization proceeds via the eigenvector matrix $\mathbf{V} = [\mathbf{v}_+, \mathbf{v}_-]$ of~$\mathbf{A}$, where $\mathbf{v}_\pm = [1, \lambda_\pm]^\top$. The matrix of eigenvectors and its inverse are:
\begin{equation}
    \mathbf{V} = \begin{pmatrix} 1 & 1 \\ \lambda_+ & \lambda_- \end{pmatrix}, \quad
    \mathbf{V}^{-1} = \frac{1}{\lambda_+ - \lambda_-}\begin{pmatrix} \lambda_- & -1 \\ -\lambda_+ & 1 \end{pmatrix}.
    \label{eq:eigenvectors}
\end{equation}
In the underdamped regime, $\lambda_+ - \lambda_- = 2i\omega_d$, so $\mathbf{V}^{-1}$ involves division by $2i\omega_d$. The complex coordinate $z = q + ip$ corresponds to a specific linear combination of the eigenbasis coordinates, chosen to preserve the physical interpretation: $q = \text{Re}(z)$ recovers the position and $p = \text{Im}(z)$ recovers the momentum. This is not an arbitrary choice; it is the unique (up to conjugation) linear transformation that simultaneously diagonalizes the dynamics and preserves the physical meaning of both components.

\paragraph{Geometric interpretation in the complex plane.}
The dynamics of Eq.~\eqref{eq:complex_ode} admit a transparent geometric interpretation. The free response ($u = 0$) with initial condition $z_0 = q_0 + ip_0$ evolves as:
\begin{equation}
    z(t) = z_0 \, e^{(-\nu + i\omega)t} = |z_0| \, e^{-\nu t} \, e^{i(\omega t + \phi_0)},
    \label{eq:free_response}
\end{equation}
where $\phi_0 = \arg(z_0)$ is the initial phase. This describes a logarithmic spiral in the complex plane: the state rotates at angular velocity $\omega$ (corresponding to the oscillator's natural frequency) while its amplitude decays exponentially at rate $\nu$. The spiral's pitch angle $\alpha = \arctan(\nu/\omega)$ determines the ratio of decay to rotation per cycle; small $\alpha$ (low damping) produces tightly wound spirals with many oscillations before decay, while large $\alpha$ (high damping) produces rapidly collapsing spirals.

For the driven system, the Green's function (impulse response) is $h(t) = e^{(-\nu+i\omega)t}\Theta(t)$ where $\Theta$ is the Heaviside step function, and the full response is the convolution $z(t) = \int_0^t h(t-s)u(s)\,ds$. This convolution structure is what enables the parallel scan formulation: it expresses the output as an associative cumulative operation over the input sequence.

\subsection{Frequency Response and Transfer Function Analysis}

The signal processing interpretation of the oscillator provides insight into what the network computes. Taking the Laplace transform of the driven oscillator $\ddot{q} + \nu\dot{q} + kq = u(t)$ yields the transfer function:
\begin{equation}
    G(s) = \frac{Q(s)}{U(s)} = \frac{1}{s^2 + \nu s + k},
    \label{eq:transfer}
\end{equation}
which is a second-order bandpass filter. Substituting $s = j\Omega$ (where $\Omega$ is the spatial frequency along a row or column), the magnitude response is
\begin{equation}
    |G(j\Omega)| = \frac{1}{\sqrt{(k - \Omega^2)^2 + \nu^2\Omega^2}},
    \label{eq:freq_response}
\end{equation}
with peak at $\Omega = \omega = \sqrt{k}$, peak magnitude $|G(j\omega)| = 1/(\nu\omega)$, and quality factor $Q = \omega/\nu$. The quality factor controls the selectivity of the filter: high $Q$ (low damping) produces a narrow bandpass that responds only to features near frequency $\omega$, while low $Q$ (high damping) produces a broad response that passes a wide range of frequencies.

A bank of $D$ oscillators with different learned stiffnesses $k_c$ forms a neural filterbank that decomposes the input into $D$ frequency channels, each with center frequency $\omega_c = \sqrt{k_c}$ and bandwidth $\nu_c$. This is structurally analogous to the cochlear filterbank in auditory processing~\cite{lyon2017human} or a bank of Gabor filters in image analysis, but with three key differences. First, the center frequencies are learned end-to-end from the task objective rather than fixed on a predefined scale. Second, the damping is input-dependent, allowing the filterbank to dynamically adjust its selectivity per spatial location. Third, each channel produces not only the filtered output (position $q_c$) but also the instantaneous spatial derivative (momentum $p_c$) and the local energy ($\mathcal{H}_c$) as structured byproducts.

\paragraph{Momentum as a spatial derivative operator.}
The momentum transfer function provides formal justification for interpreting $p$ as a boundary detector. Taking the Laplace transform of $p = \dot{q}$ yields $P(s) = sQ(s) - q(0)$, so the momentum transfer function is:
\begin{equation}
    G_p(s) = \frac{P(s)}{U(s)} = \frac{s}{s^2 + \nu s + k},
    \label{eq:momentum_transfer}
\end{equation}
which is a \emph{bandpass differentiator}: the numerator $s$ corresponds to spatial differentiation, while the denominator provides frequency selectivity. At frequencies near the resonance $\Omega \approx \omega$, the momentum response peaks at $|G_p(j\omega)| = 1/\nu$, exactly $Q$ times larger than the position response at the same frequency. This means that the momentum is an amplified, frequency-selective spatial derivative, precisely the operation needed to detect boundaries at a particular spatial scale. The factor-of-$Q$ amplification explains why momentum carries a stronger boundary signal than naive finite differences: the oscillator accumulates derivative information over a spatial window of length $\sim Q/\omega$, effectively averaging out noise while preserving sharp transitions.

\paragraph{Energy as a scale-space saliency measure.}
The energy $\mathcal{H}_c = \frac{1}{2}(q_c^2 + p_c^2)$ combines both position and momentum into a single non-negative scalar field. In the frequency domain, the energy spectral density satisfies:
\begin{equation}
    S_{\mathcal{H}}(\Omega) = \tfrac{1}{2}\big(|G(j\Omega)|^2 + |G_p(j\Omega)|^2\big) \cdot |U(j\Omega)|^2 = \frac{1 + \Omega^2}{2\big[(k - \Omega^2)^2 + \nu^2\Omega^2\big]} \cdot |U(j\Omega)|^2.
    \label{eq:energy_spectral}
\end{equation}
This expression reveals that energy responds to a broader spectral band than either position or momentum alone, since it integrates both the low-frequency content (via $|G|^2$) and the high-frequency derivative content (via $|G_p|^2$). The resulting energy map therefore functions as a saliency detector that highlights regions where the input signal has significant power in the neighborhood of the oscillator's resonant frequency, regardless of whether that power comes from slowly varying features or sharp transitions. This dual sensitivity is what makes energy suitable for gating skip connections: salient regions are highlighted whether they contain smooth interior texture or sharp boundary gradients.

\paragraph{Parseval's relation and energy conservation.}
The relationship between time-domain energy and frequency-domain content provides a useful identity. For a finite-length signal of length $L$ driven through a single oscillator, the total accumulated energy satisfies:
\begin{equation}
    \sum_{t=0}^{L-1} \mathcal{H}_c(t) = \frac{1}{2}\sum_{t=0}^{L-1}\big(q_c(t)^2 + p_c(t)^2\big) = \frac{1}{2L}\sum_{n=0}^{L-1}\big(|G(j\Omega_n)|^2 + |G_p(j\Omega_n)|^2\big)|U(j\Omega_n)|^2,
    \label{eq:parseval}
\end{equation}
where $\Omega_n = 2\pi n/L$ are the discrete frequencies. This is a generalized Parseval relation for the oscillator, establishing that the spatial average of energy is a weighted frequency-domain integral of the input power spectrum. For classification, where we globally pool energy via $\bar{h} = \text{GAP}(H_\text{map})$, this identity shows that the pooled energy statistic is literally a nonlinear spectral summary of the input, with the learned oscillator frequencies $\omega_c$ determining which spectral components contribute most.

\subsection{Discretization and Efficient Implementation}

For discrete inputs at spatial positions $t = 0, 1, \ldots, L{-}1$ along a row or column, we discretize Eq.~\eqref{eq:complex_ode} with input-dependent step size $\Delta_t$ and damping $\nu_t$:
\begin{equation}
    z_t = \underbrace{\exp(-\nu_t\Delta_t)}_{\text{decay}\; |\bar{A}_t| < 1} \cdot \underbrace{e^{i\omega\Delta_t}}_{\text{rotation}} \cdot z_{t-1} + u_t,
    \label{eq:recurrence}
\end{equation}
where the input-dependent parameters are computed as $\nu_t = \text{clamp}(\text{softplus}(x_t \odot s_\nu + b_\nu),\; \max{=}5)$ and $\Delta_t = \text{softplus}(x_t \odot s_\Delta + b_\Delta)$, with $s_\nu, b_\nu, s_\Delta, b_\Delta$ learnable per channel. The discrete transition coefficient $\bar{A}_t = \exp(-\nu_t\Delta_t + i\omega\Delta_t)$ lies on the damped unit circle in the complex plane, with $|\bar{A}_t| = \exp(-\nu_t\Delta_t) < 1$ by construction. This guarantees that $|z_t| \leq |z_{t-1}| + |u_t|$ at every step, a bounded-input bounded-output (BIBO) stability property that prevents hidden state explosion without any normalization layers.

This recurrence has the general form $z_t = \bar{A}_t z_{t-1} + u_t$ of a linear state-space model. In generic SSMs such as Mamba~\cite{gu2023mamba}, $\bar{A}$ is an arbitrary learned diagonal with no structural constraint. Table~\ref{tab:ssm_comparison} summarizes the distinctions.

\begin{table}[t]
\centering
\caption{Generic state-space models versus the Hamiltonian oscillator. The physics constraint reduces the parameter space but enriches the representation with structured outputs.}
\label{tab:ssm_comparison}
\small
\begin{tabular}{@{}lcc@{}}
\toprule
Property & Generic SSM & Hamiltonian (ours) \\
\midrule
Transition $\bar{A}$ & Learned diagonal & $e^{(-\nu+i\omega)\Delta}$ \\
State space & Real $\mathbb{R}^N$ & Complex $\mathbb{C}^N$ \\
$|\bar{A}|$ bounded? & Not guaranteed & $< 1$ by construction \\
Frequency structure & None & $\omega_c$ per channel \\
Quality factor & None & $Q_c = \omega_c/\nu_c$ \\
Outputs & Features only & $q$, $p$, $\mathcal{H}$ \\
Stability guarantee & None & BIBO (Eq.~\ref{eq:energy_rate}) \\
\bottomrule
\end{tabular}
\end{table}

The recurrence~\eqref{eq:recurrence} can be unrolled into a closed-form expression. Defining the cumulative log-coefficient $L_t = \sum_{j=0}^{t}(-\nu_j\Delta_j + i\omega\Delta_j)$, the state at position $t$ is:
\begin{equation}
    z_t = e^{L_t} \sum_{j=0}^{t} e^{-L_j} u_j.
    \label{eq:parallel_scan}
\end{equation}
This formulation enables parallel computation via cumulative sums in $O(\log L)$ parallel depth. However, for 2D feature maps flattened into a single sequence of length $H \times W$, the cumulative decay $|L_t|$ can reach extreme values. At the bottleneck resolution ($28 \times 28$), a flattened scan of length 784 would produce $|L_{784}| \approx 400$, and $\exp(400)$ overflows float32. We resolve this by scanning rows and columns independently in four directions (left-to-right, right-to-left, top-to-bottom, bottom-to-top), each processing sequences of length 28. The maximum cumulative magnitude is then approximately 14, yielding $\exp(14) \approx 1.2 \times 10^6$, well within float32 range. This is not merely a numerical workaround; it is physically natural since each row and column constitutes an independent oscillator line, and the four directions capture complementary horizontal and vertical spatial information. The four direction-specific outputs are merged via learned linear projections:
\begin{equation}
    q_\text{merged} = W_q [q^{\rightarrow}; q^{\leftarrow}; q^{\downarrow}; q^{\uparrow}], \quad p_\text{merged} = W_p [p^{\rightarrow}; p^{\leftarrow}; p^{\downarrow}; p^{\uparrow}].
    \label{eq:direction_merge}
\end{equation}

\paragraph{Discretization error and the exponential integrator.}
The discretization in Eq.~\eqref{eq:recurrence} is an \emph{exponential-Euler} method, which integrates the homogeneous part $\dot{z} = (-\nu+i\omega)z$ exactly (via the matrix exponential $e^{(-\nu+i\omega)\Delta}$) and approximates the forcing as piecewise constant over each step. The local truncation error is $O(\Delta^2)$ for the forcing contribution, but critically, the exponential factor is exact regardless of step size. This means that the oscillatory and dissipative dynamics are captured without error, only the abruptness of input transitions introduces discretization artifacts. This is a significant advantage over the forward-Euler discretization $z_t = (1 + (-\nu+i\omega)\Delta)z_{t-1} + \Delta u_t$ used in some SSMs, which introduces a spurious amplification of $|1 + (-\nu+i\omega)\Delta|$ that can exceed 1 for large $\Delta$, breaking the BIBO stability guarantee. With the exponential integrator, $|\bar{A}_t| = e^{-\nu_t \Delta_t} < 1$ unconditionally, preserving stability for any positive step size.

\paragraph{Effective receptive field of the oscillator.}
The impulse response $h_t = \bar{A}_0 \bar{A}_1 \cdots \bar{A}_{t-1}$ of the discrete oscillator decays as $|h_t| = \exp(-\sum_{j=0}^{t-1}\nu_j\Delta_j)$. For constant parameters, this simplifies to $|h_t| = e^{-\nu\Delta t}$, giving an effective memory length of $\tau_\text{eff} = 1/(\nu\Delta)$ steps. At the bottleneck resolution ($L = 28$ per row), a channel with $\nu\Delta \approx 0.1$ has $\tau_\text{eff} \approx 10$ pixels, while $\nu\Delta \approx 0.01$ yields $\tau_\text{eff} \approx 100$ pixels, far exceeding the row length. The input-dependence of $\nu_t$ and $\Delta_t$ allows the network to dynamically adjust this receptive field: in homogeneous regions, larger $\Delta_t$ produces longer memory (integrating context), while at boundaries, smaller $\Delta_t$ shortens the memory (preserving sharp transitions). This adaptive receptive field is a structural advantage over fixed-kernel convolutions.

\paragraph{Computational complexity.}
The parallel scan over a sequence of length $L$ requires $O(L)$ work and $O(\log L)$ parallel depth. For $D = 384$ channels scanned in four directions over rows and columns of length $L = 28$, the total work per feature map is $4 \times D \times O(L) = O(4DL)$, which at our operating point amounts to $4 \times 384 \times 28 \approx 43{,}000$ complex multiply-adds. By comparison, the ConvNeXt block in the same bottleneck involves a $7 \times 7$ depthwise convolution ($384 \times 49 \times 784 \approx 14.8\text{M}$ multiply-adds) followed by two pointwise convolutions ($384 \times 1536 \times 784 \approx 463\text{M}$), so the oscillator bank adds less than 0.01\% overhead relative to the ConvNeXt path. The dominant cost remains the ConvNeXt blocks in the encoder and decoder; the oscillator bank is computationally negligible.

\subsection{Shared Encoder and Hamiltonian Bottleneck}

We refer to the overall architecture as \textbf{HamVision}, encompassing the shared encoder and Hamiltonian bottleneck together with task-specific heads: HamSeg for segmentation and HamCls for classification. Both configurations share an identical encoder and bottleneck. The encoder comprises a two-layer convolutional stem followed by three stages of ConvNeXt blocks~\cite{liu2022convnext} with spatial downsampling, where channels progress as $C \to 2C \to 4C \to 8C$ with $C = 48$. At the deepest level ($8C = 384$ channels, $28 \times 28$), the Hamiltonian bottleneck processes features through two parallel paths: a ConvNeXt block providing reliable gradient flow, and the oscillator bank providing physics-structured processing. A learned gate $g = \sigma(W[f_\text{conv}; f_\text{osc}])$ fuses the two paths, with bias initialized to $+2.0$ to favor ConvNeXt early in training. Dropout ($p{=}0.1$) regularizes the fused output.

The per-channel energy values $\mathcal{H}_c(i,j) = \tfrac{1}{2}(q_c(i,j)^2 + p_c(i,j)^2)$ form a $D$-channel energy tensor. Not all channels contribute equally to boundary detection; some oscillator frequencies may respond to irrelevant features. We collapse this tensor into a single-channel spatial map using a squeeze-and-excitation attention mechanism~\cite{hu2018senet} that learns to weight channels by their discriminative relevance:
\begin{equation}
    H_\text{map}(i,j) = \frac{1}{D}\sum_{c=1}^{D} w_c \cdot \mathcal{H}_c(i,j), \quad \bm{w} = \sigma\!\big(\text{MLP}(\text{GAP}(\mathcal{H}))\big).
    \label{eq:energy_map}
\end{equation}

The bottleneck produces three structured outputs: fused features $f$ (for task-specific processing), momentum tensor $p \in \mathbb{R}^{D \times H \times W}$ (spatial gradients), and energy map $H_\text{map} \in \mathbb{R}^{1 \times H \times W}$ (spatial saliency).

\paragraph{Gate dynamics and training stability.}
The gate initialization $b_g = +2.0$ ensures that $\sigma(W[\cdot] + 2.0) \approx 0.88$ at the start of training, meaning the bottleneck initially behaves almost identically to a pure ConvNeXt block. This warm-start strategy addresses a practical challenge: the oscillator bank produces meaningful outputs only after its frequency and damping parameters have been calibrated, which requires gradient signals from the task loss. If the gate started at 0.5, the oscillator's initially random outputs would corrupt the ConvNeXt features, destabilizing early training. The bias decays during training as the oscillator becomes useful; empirically, the gate converges to $\bar{g} \approx 0.54$ (Table~\ref{tab:diagnostics}), indicating that the network learns to allocate roughly equal weight to both paths.

\paragraph{Information-theoretic interpretation of the SE attention.}
The squeeze-and-excitation module that produces $H_\text{map}$ from the per-channel energy tensor can be understood through an information-theoretic lens. Each oscillator channel $c$ responds to a particular frequency band centered at $\omega_c$; the energy $\mathcal{H}_c(i,j)$ measures how much power the input signal has in that band at spatial location $(i,j)$. Not all frequency bands are equally informative for the task; the SE weights $w_c$ learn to suppress irrelevant bands (noise, background texture) and amplify discriminative ones (boundary frequencies, pathological patterns). The resulting map $H_\text{map} = \sum_c w_c \mathcal{H}_c / D$ is a learned spectral projection: a weighted sum over frequency channels that extracts the single most task-relevant saliency signal from the full spectral decomposition.

\begin{figure}[H]
    \centering
    \includegraphics[width=1\linewidth]{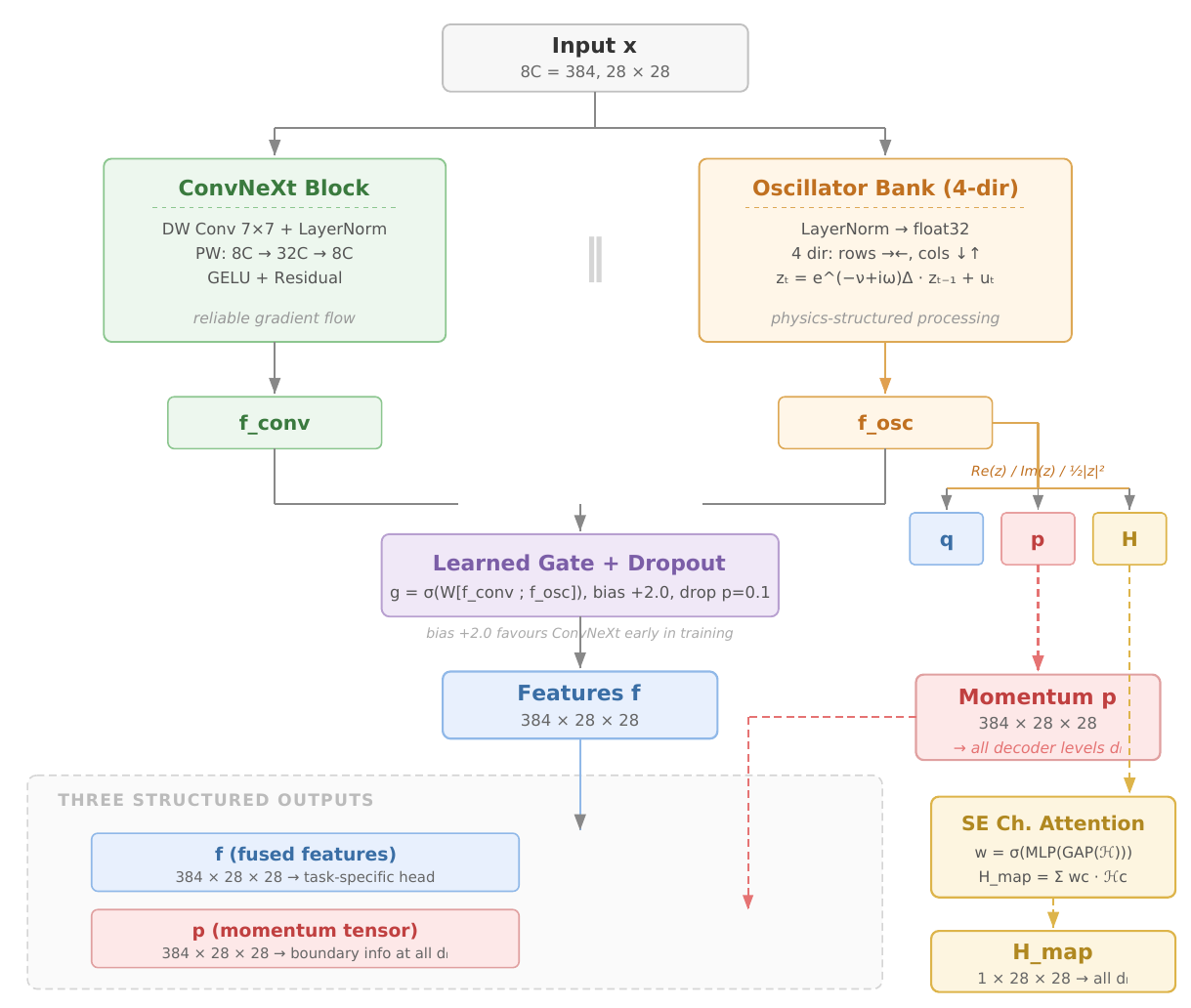}
    \caption{Detailed architecture of the Hamiltonian Bottleneck. The input is processed through two parallel paths: a ConvNeXt block (left) providing reliable gradient flow, and an oscillator bank (right) performing physics-structured processing via four-directional scanning. The oscillator's complex state~$z$ is decomposed into position~$q$, momentum~$p$, and energy~$H = \tfrac{1}{2}|z|^2$. A learned gate fuses the ConvNeXt and oscillator outputs into features~$f$, while a squeeze-and-excitation channel attention module collapses the per-channel energy tensor into a single-channel saliency map~$H_\text{map}$. All three outputs propagate to downstream task-specific heads.}
    \label{fig:hamblock_detail}
\end{figure}

\subsection{HamSeg: Segmentation Head}

For segmentation, the decoder mirrors the encoder with transposed convolutions for upsampling and ConvNeXt blocks at each level. At every decoder stage $l \in \{1, 2, 3\}$, the oscillator's outputs modulate the skip connections through energy gating and momentum injection.

The energy map, interpolated to the spatial resolution of decoder level $l$ and centered around its spatial mean, gates the encoder features:
\begin{equation}
    \tilde{e}_l = \sigma\!\big(\gamma_l \cdot (H_l - \bar{H}_l)\big) \odot e_l,
    \label{eq:gate}
\end{equation}
where $\gamma_l$ is a learned scalar and $\bar{H}_l = \frac{1}{HW}\sum_{i,j}H_l(i,j)$ is the spatial mean. The centering is essential: without it, the sigmoid receives raw energy values that are typically large and positive, saturating near 1.0 and rendering the gate useless. With centering, the gate operates around $\sigma(0) = 0.5$, and the learned gain $\gamma_l$ controls the contrast, regions with above-average excitation (boundaries, salient structures) are amplified while below-average regions (homogeneous background) are suppressed.

Simultaneously, the oscillator momentum is projected through a $1 \times 1$ convolution to match the channel dimension of decoder level $l$ and concatenated with the decoder and gated encoder features:
\begin{equation}
    d_l = \text{CNX}\!\big([\tilde{e}_l;\; d_{l+1}^{\uparrow};\; W_l^{(p)} p_l]\big),
    \label{eq:decoder_fusion}
\end{equation}
where $d_{l+1}^{\uparrow}$ denotes the upsampled output from the previous decoder stage, $W_l^{(p)} \in \mathbb{R}^{C_l \times D}$ is a $1 \times 1$ projection, $p_l$ is the momentum tensor interpolated to the spatial resolution of level $l$, and $\text{CNX}(\cdot)$ denotes a ConvNeXt block. The concatenated tensor has $3C_l$ channels (gated encoder, upsampled decoder, projected momentum), and the ConvNeXt block learns to fuse boundary information from the oscillator with spatial features from the encoder and the decoder. Momentum injection at all three decoder levels, not just the coarsest, was the single most impactful design choice, contributing $+5.59\%$ Dice on ACDC where multi-scale cardiac structures demand boundary information at every resolution.

\paragraph{Why centering is necessary for the energy gate.}
The energy values $H_l(i,j)$ are inherently non-negative and typically large in magnitude (empirically, $\bar{H} \approx 5$--$15$ depending on the input). Applying $\sigma(H_l)$ directly would saturate above 0.99 everywhere, producing a near-constant gate. Centering by the spatial mean transforms the gate argument into a zero-mean quantity: $\sigma(\gamma_l(H_l - \bar{H}_l))$ operates around the sigmoid's sensitive region near $\sigma(0) = 0.5$, with $\gamma_l$ controlling the contrast. The centered gate can be interpreted as a soft spatial selection operator: locations where the local energy exceeds the image-wide average are amplified (gate $> 0.5$), while locations with below-average energy are suppressed (gate $< 0.5$). This centering is both physically motivated (absolute energy depends on input scale, but relative energy is invariant) and practically essential (without it, the gate collapses to a constant).

\paragraph{Phase-space attention at the coarsest level.}
At decoder stage $d_3$ (the coarsest), a phase-space attention mechanism provides additional modulation. This mechanism computes spatial attention weights from centered energy and combines them with projected momentum:
\begin{equation}
    \alpha(i,j) = \sigma\!\big(\gamma_\text{ps}(H(i,j) - \bar{H})\big), \quad
    a_l = \alpha \odot W_\text{ps}^{(p)} p + d_l,
    \label{eq:ps_attention}
\end{equation}
where $W_\text{ps}^{(p)}$ is a $1 \times 1$ projection and the attention-modulated momentum is added as a residual to the decoder features. The attention weights $\alpha$ use the same centering strategy as the skip gates but with an independent gain $\gamma_\text{ps}$, allowing the network to learn a different contrast level for the coarsest decoder stage where spatial selection is most critical. The multiplication $\alpha \odot W_\text{ps}^{(p)} p$ has a clear physical interpretation: it selectively injects boundary information (momentum) only at spatially salient locations (high energy), preventing the decoder from being distracted by momentum signals in irrelevant background regions.

A segmentation head ($3 \times 3$ convolution followed by $1 \times 1$ projection) produces the final prediction. The segmentation network totals 8.57M parameters.
\begin{figure}[H]
    \centering
    \includegraphics[width=1\linewidth]{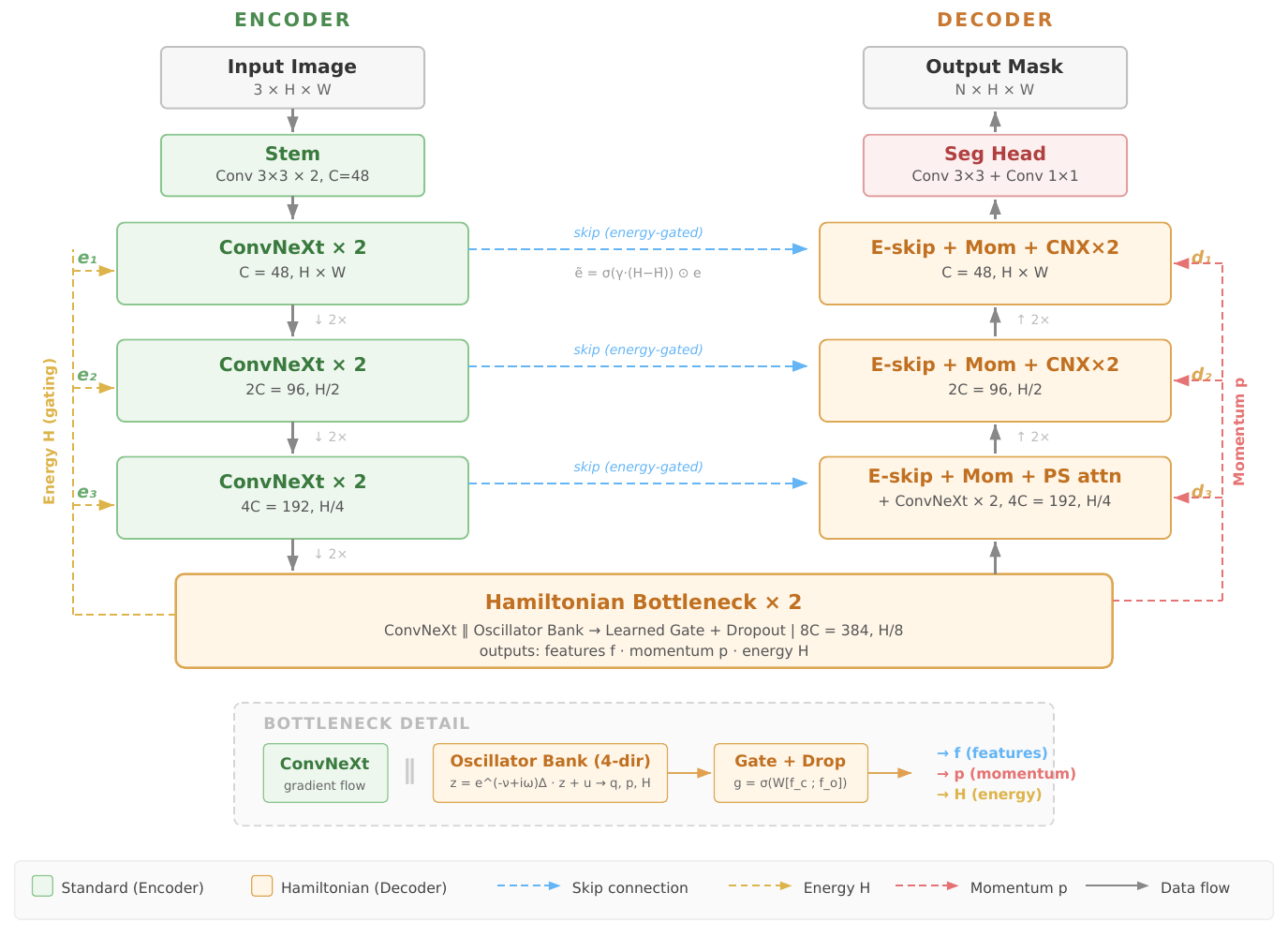}
    \caption{Overview of the HamSeg segmentation architecture. A shared ConvNeXt encoder with a Hamiltonian oscillator bottleneck produces position~$q$, momentum~$p$, and energy~$H$ representations, which are injected into a U-Net decoder via energy-gated skip connections and multi-scale momentum concatenation.}
    \label{fig:hamseg_arch}
\end{figure}

\subsection{HamCls: Classification Head}

For classification, the U-Net decoder is replaced by a phase-space pooling head that aggregates the oscillator's three outputs into a compact classification vector. The key insight is that the same physical quantities that provide spatial information for segmentation provide statistical information for classification when globally pooled.

Features are globally averaged into a $D$-dimensional vector $\bar{f} = \text{GAP}(f) \in \mathbb{R}^{384}$, encoding what content the image contains. Momentum is globally averaged into a $D$-dimensional vector $\bar{p} = \text{GAP}(|p|) \in \mathbb{R}^{384}$, encoding how spatially active the image is, a measure of texture complexity and boundary density. The energy map is pooled and passed through a small MLP to produce a low-dimensional statistics vector $\bar{h} = \text{MLP}(\text{GAP}(H_\text{map})) \in \mathbb{R}^{16}$, encoding the overall excitation level.

These three vectors are concatenated into a 784-dimensional phase-space representation:
\begin{equation}
    \mathbf{v}_\text{ps} = [\bar{f};\; \bar{p};\; \bar{h}] \in \mathbb{R}^{784},
    \label{eq:ps_vector}
\end{equation}
which is passed through LayerNorm and classified by a two-layer MLP with GELU activation and dropout:
\begin{equation}
    \hat{y} = W_2 \, \text{GELU}\!\big(\text{Drop}(W_1 \, \text{LN}(\mathbf{v}_\text{ps}) + b_1)\big) + b_2,
    \label{eq:cls_head}
\end{equation}
where $W_1 \in \mathbb{R}^{384 \times 784}$, $W_2 \in \mathbb{R}^{K \times 384}$, and $K$ is the number of classes. The classification network totals approximately 7.3M parameters.

\paragraph{Physical semantics of each pooled component.}
The three components of $\mathbf{v}_\text{ps}$ capture complementary aspects of the image, grounded in the oscillator's physical interpretation. The feature vector $\bar{f}$ summarizes \emph{what} content the image contains, analogous to pooling a standard CNN backbone. The momentum vector $\bar{p}$ summarizes \emph{how spatially active} the image is across frequency channels: diseases with complex textures (e.g., irregular cell morphologies in BloodMNIST) produce high average momentum, while smooth pathologies produce low momentum. The energy statistic $\bar{h}$ summarizes the \emph{overall excitation profile}: by compressing the 1-channel energy map through GAP and a small MLP into 16 dimensions, the network learns task-relevant nonlinear statistics of the energy distribution (mean, variance, skewness-like features) without explicit supervision.

\paragraph{Why the three components are not redundant.}
A natural question is whether $\bar{p}$ and $\bar{h}$ provide information beyond what $\bar{f}$ already captures. The answer lies in the algebraic relationships. Features $f$ are the output of the gated fusion, which mixes ConvNeXt and oscillator paths: $f = g \odot f_\text{conv} + (1-g) \odot f_\text{osc}$. Momentum $p$ is extracted directly from the oscillator's imaginary component \emph{before} gating, so it preserves raw boundary information that gating may attenuate. Energy $H_\text{map}$ is a nonlinear function $\frac{1}{2}(q^2 + p^2)$ that creates cross-terms between position and momentum unavailable in either alone. Moreover, GAP$(|p|)$ computes the $L^1$ norm of the momentum field (a measure of total variation), while GAP$(f)$ computes the mean feature activation. These are fundamentally different statistics of the spatial field: the first measures how much the oscillator state \emph{changes} across space, while the second measures what the state \emph{is} on average.

The oscillator bottleneck is identical between HamSeg and HamCls; only the downstream head differs. Together, these two configurations constitute the HamVision framework.

\begin{figure}[H]
    \centering
    \includegraphics[width=1\linewidth]{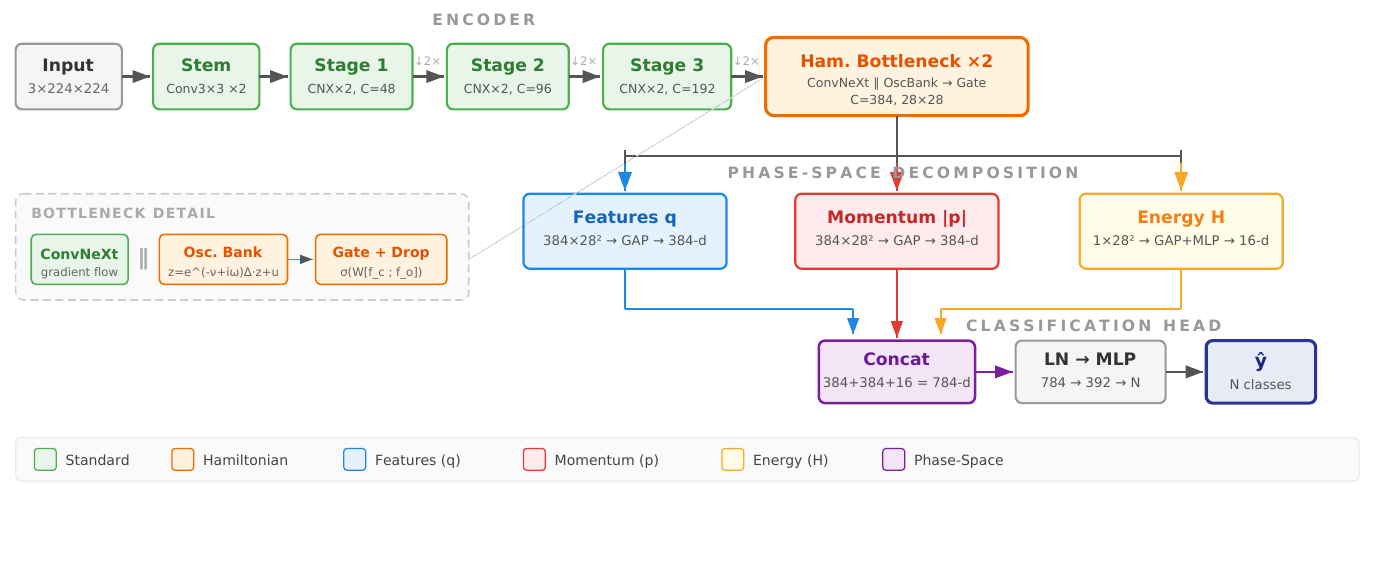}
    \caption{Architecture of HamCls, the classification head of HamVision. The shared encoder (Stem $\to$ ConvNeXt stages $\to$ Hamiltonian Bottleneck) produces three phase-space quantities: features~$q$, momentum magnitude~$|p|$, and scalar energy~$H$. Each is globally pooled and concatenated into a 784-dimensional phase-space vector, which is classified by a LayerNorm--MLP head.}
    \label{fig:hamcls_arch}
\end{figure}

\section{Experiments}
\label{sec:experiments}

We evaluate HamVision on two tasks: segmentation (four benchmarks, three modalities) and classification (six MedMNIST benchmarks, five modalities), for a total of ten datasets.

\subsection{Segmentation}

We evaluate on ISIC\,2018~\cite{codella2019isic2018} (2,694 dermoscopic images, binary), ISIC\,2017 (2,150 dermoscopic images, binary), TN3K (3,493 thyroid ultrasound images, binary), and ACDC~\cite{bernard2018acdc} (100 cardiac MRI patients, 4-class). Training uses AdamW with learning rate $5 \times 10^{-4}$, cosine annealing, 200 epochs, Dice+BCE loss (binary) or Dice+CE (multi-class), and $224^2$ input with standard augmentation.

\begin{table}[t]
\centering
\caption{Segmentation results. Best in \textbf{bold}, second \underline{underlined}. HamSeg uses 8.57M parameters throughout.}
\label{tab:seg_results}
\small
\begin{tabular}{@{}lccc@{}}
\toprule
Method & Params & DSC & mIoU \\
\midrule
\multicolumn{4}{@{}l}{\textit{ISIC\,2018 (dermoscopy, binary)}} \\
U-Net & -- & 87.06 & 79.08 \\
TransUNet & ${\sim}$105M & 87.47 & 79.50 \\
SwinUNet & ${\sim}$27M & 87.16 & 79.54 \\
VM-UNet & ${\sim}$36M & 88.10 & 78.89 \\
FreqConvMamba & ${\sim}$16M & \underline{89.19} & \underline{80.64} \\
HamSeg & \textbf{8.57M} & \textbf{89.38} & \textbf{81.22} \\
\midrule
\multicolumn{4}{@{}l}{\textit{ISIC\,2017 (dermoscopy, binary)}} \\
TransUNet & ${\sim}$105M & 83.27 & 73.54 \\
VM-UNet & ${\sim}$36M & 87.42 & -- \\
FreqConvMamba & ${\sim}$16M & \underline{88.37} & \underline{79.01} \\
HamSeg & \textbf{8.57M} & \textbf{88.40} & \textbf{79.20} \\
\midrule
\multicolumn{4}{@{}l}{\textit{TN3K (thyroid ultrasound, binary)}} \\
TransUNet & ${\sim}$105M & 80.89 & 71.23 \\
VM-UNet & ${\sim}$36M & 83.65 & 71.90 \\
FreqConvMamba & ${\sim}$16M & \underline{86.85} & \underline{76.76} \\
HamSeg & \textbf{8.57M} & \textbf{87.05} & \textbf{76.33} \\
\midrule
\multicolumn{4}{@{}l}{\textit{ACDC (cardiac MRI, 4-class)}} \\
TransUNet & ${\sim}$105M & 87.43 & -- \\
FreqConvMamba & ${\sim}$16M & \underline{89.79} & -- \\
HamSeg & \textbf{8.57M} & \textbf{92.40} & \textbf{86.14} \\
\bottomrule
\end{tabular}
\end{table}

Table~\ref{tab:seg_results} summarizes the results. HamSeg achieves the highest Dice score on all four benchmarks with 8.57M parameters, 2--12$\times$ fewer than competing methods. The most striking result is on ACDC ($+2.61\%$ over FreqConvMamba), where the three cardiac structures vary enormously in size, placing a premium on boundary information at multiple resolutions. The consistency across dermoscopy, ultrasound, and MRI suggests that the oscillator's inductive bias captures something fundamental about the segmentation task rather than properties specific to any imaging modality.

\begin{figure}[H]
    \centering
    \includegraphics[width=0.7\linewidth]{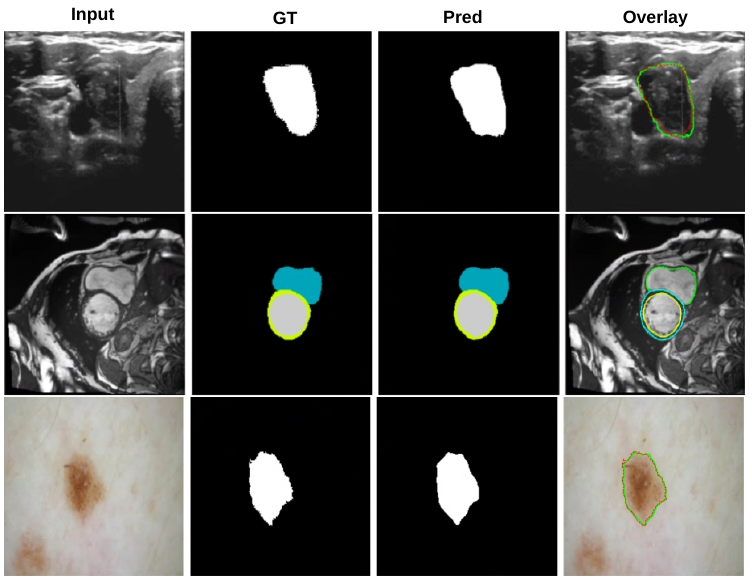}
    \caption{Qualitative segmentation for different datasets (TN3K thyroid ultrasound, ACDC cardiac MRI, and ISIC\,2018 dermoscopy). Left to right: input, ground truth, predicted segmentation, and overlay.}
    \label{fig:placeholder}
\end{figure}

\subsection{Classification}

We evaluate on six MedMNIST datasets~\cite{yang2023medmnist} at $224 \times 224$ resolution: PathMNIST (colon pathology, 9 classes), DermaMNIST (dermatoscopy, 7 classes), BloodMNIST (blood cell microscopy, 8 classes), BreastMNIST (breast ultrasound, 2 classes), RetinaMNIST (retinal fundus, 5 classes), and OrganSMNIST (abdominal CT sagittal, 11 classes). All use the official MedMNIST splits. Training uses AdamW with cosine annealing, $224^2$ input, and standard augmentation. For imbalanced datasets (BreastMNIST), inverse-frequency class weights are applied.

\begin{table}[t]
\centering
\caption{Classification accuracy (\%) on six MedMNIST benchmarks ($224{\times}224$). Best in \textbf{bold}, second \underline{underlined}. Baselines from~\cite{yue2024medmamba}.}
\label{tab:cls_results}
\footnotesize
\setlength{\tabcolsep}{3pt}
\begin{tabular}{@{}llcc@{}}
\toprule
Dataset & Method & AUC & ACC \\
\midrule
\multirow{3}{*}{BloodMNIST}
  & ResNet-50   & 99.7 & 95.6 \\
  & MedMamba-S  & \underline{99.9} & \underline{98.4} \\
  & \textbf{HamCls} & \textbf{99.93} & \textbf{98.85} \\
\midrule
\multirow{3}{*}{PathMNIST}
  & MedViT-S    & 99.3 & 94.2 \\
  & MedMamba-B  & \underline{99.9} & \underline{96.4} \\
  & \textbf{HamCls} & 99.36 & \textbf{96.65} \\
\midrule
\multirow{3}{*}{DermaMNIST}
  & MedViT-S    & \underline{93.7} & \textbf{78.0} \\
  & MedMamba-T  & 91.7 & 77.9 \\
  & \textbf{HamCls} & \textbf{93.66} & \underline{77.96} \\
\midrule
\multirow{3}{*}{BreastMNIST}
  & MedViT-S    & \textbf{93.8} & \textbf{89.7} \\
  & MedMamba-B  & 84.9 & \underline{89.1} \\
  & \textbf{HamCls} & \underline{89.94} & 89.60 \\
\midrule
\multirow{3}{*}{OrganSMNIST}
  & MedViT-S    & \underline{98.7} & 80.5 \\
  & MedMamba-B  & 98.4 & \textbf{83.4} \\
  & \textbf{HamCls} & \textbf{98.02} & \underline{80.96} \\
\midrule
\multirow{3}{*}{RetinaMNIST}
  & MedViT-S    & \underline{75.2} & 53.4 \\
  & MedMamba-X  & 71.9 & \underline{57.0} \\
  & \textbf{HamCls} & \textbf{76.24} & \textbf{56.75} \\
\bottomrule
\end{tabular}
\end{table}

Table~\ref{tab:cls_results} presents results across six MedMNIST benchmarks. On BloodMNIST, HamCls achieves 98.85\% accuracy, exceeding MedMamba-S by $+0.45\%$. On PathMNIST, HamCls achieves 96.65\% accuracy, exceeding MedMamba-B by $+0.25\%$. On DermaMNIST, HamCls matches MedViT-S on AUC (93.66\% vs.\ 93.7\%) while maintaining competitive accuracy at 77.96\%. On BreastMNIST, which contains only 780 images and severe class imbalance, HamCls achieves 89.60\% accuracy, approaching MedViT-S (89.7\%). On OrganSMNIST and RetinaMNIST, HamCls obtains competitive results, demonstrating consistency across datasets with diverse characteristics, image statistics, and anatomical focus.

\subsection{Interpretability and Signal Analysis}
\label{sec:vis}

A central advantage of the Hamiltonian formulation is that every intermediate quantity carries physical meaning that can be inspected, measured, and verified without any post-hoc explanation method. Unlike attention maps (which require gradient-based attribution) or generic feature maps (which lack semantic structure), the oscillator's outputs have a priori physical interpretations that we can validate empirically. We organize this analysis around three questions: Is the oscillator actually used? Do its outputs carry the predicted physical semantics? And do these semantics differ meaningfully between tasks?

\subsubsection{Is the Oscillator Functionally Active?}

The gated fusion architecture provides a natural diagnostic: if the learned gate $g$ converges to 1.0 everywhere, the model has learned to bypass the oscillator entirely and rely only on ConvNeXt. Table~\ref{tab:diagnostics} reports measurements on the ISIC\,2018 test set. The gate mean of 0.54 with spatial standard deviation 0.043 indicates that roughly 46\% of the bottleneck output comes from the oscillator path, the model has not learned to ignore it. More revealing is the per-block breakdown: the oscillator contributes 79.6\% in the first bottleneck block and 58.8\% in the second. The higher reliance in the first block suggests that the oscillator performs initial frequency decomposition, while the second block allows the ConvNeXt path to refine the representation with conventional spatial processing.

The phase-space attention produces energy-based modulation weights spanning $[0.16, 0.78]$ with standard deviation 0.136. This full dynamic range was achieved by centering energy around its spatial mean before applying the sigmoid, without centering, raw energy values (typically $> 5$) saturate the sigmoid above 0.99, collapsing the attention to a constant and rendering it useless. This design choice, motivated by the physics (energy is an extensive quantity whose absolute magnitude is less informative than its spatial variation), proved essential for functional attention.

\begin{table}[t]
\centering
\caption{Quantitative signal diagnostics on ISIC\,2018 test samples. Every oscillator-derived signal is functionally active and produces physically interpretable measurements.}
\label{tab:diagnostics}
\small
\begin{tabular}{@{}lc@{}}
\toprule
Diagnostic measure & Value \\
\midrule
\multicolumn{2}{@{}l}{\textit{Bottleneck gate (oscillator usage)}} \\
\quad Gate $g$ mean / spatial std & 0.54 / 0.043 \\
\quad Oscillator contribution (block 0 / 1) & 79.6\% / 58.8\% \\
\midrule
\multicolumn{2}{@{}l}{\textit{Phase-space attention (energy modulation)}} \\
\quad Attention $\alpha$ range / std & [0.16, 0.78] / 0.136 \\
\midrule
\multicolumn{2}{@{}l}{\textit{Energy-gated skip connections}} \\
\quad $d_3$ gate range (learned $\gamma{=}1.18$) & [0.19, 0.75] \\
\quad $d_2$ gate range (learned $\gamma{=}0.87$) & [0.26, 0.69] \\
\quad $d_1$ gate range (learned $\gamma{=}0.91$) & [0.25, 0.70] \\
\midrule
\multicolumn{2}{@{}l}{\textit{Momentum by region (region-conditioned)}} \\
\quad Interior / Boundary / Exterior & 110.2 / 95.1 / 83.2 \\
\quad Interior / Exterior ratio & 1.33 \\
\midrule
\multicolumn{2}{@{}l}{\textit{Energy by region}} \\
\quad Boundary / Exterior ratio & 1.25 \\
\bottomrule
\end{tabular}
\end{table}

\subsubsection{Does Momentum Encode Boundary Information?}

The physics predicts that momentum $|p|$ should be large wherever the input signal changes abruptly (boundaries) and small in homogeneous regions. We test this by conditioning momentum magnitude on ground-truth regions: interior (inside the lesion mask, eroded by 5 pixels), boundary (within 5 pixels of the mask edge), and exterior (outside the mask, eroded by 5 pixels).

The measured values (Table~\ref{tab:diagnostics}) reveal a consistent gradient: interior momentum (110.2) exceeds boundary momentum (95.1), which exceeds exterior momentum (83.2). The interior-to-exterior ratio of 1.33 is substantial and consistent across test samples. This ordering might seem counterintuitive, one expects boundaries to have the highest momentum. The explanation lies in the scanning dynamics: as the oscillator scans through the lesion interior, the sustained coherent forcing (consistent feature values within the lesion) builds up momentum through resonance; at the boundary, the abrupt feature transition partially disrupts this accumulation; in the exterior, the oscillator decays toward equilibrium with low-amplitude features. The momentum thus encodes not just boundary location but also region identity, a richer signal than binary edge maps.

Figure~\ref{fig:physics_vis} visualizes this across all four segmentation datasets. The momentum maps (rightmost columns) show bright regions co-localized with the lesion interior and boundary, with darker exterior regions, consistent across dermoscopy (rows 2--3), cardiac MRI (row 1), and thyroid ultrasound (row 4). This consistency across radically different imaging modalities confirms that the momentum pattern is a property of the oscillator dynamics, not of any particular image content.

\subsubsection{Does Energy Encode Spatial Saliency?}

\begin{figure*}[t]
\centering
\includegraphics[width=\textwidth]{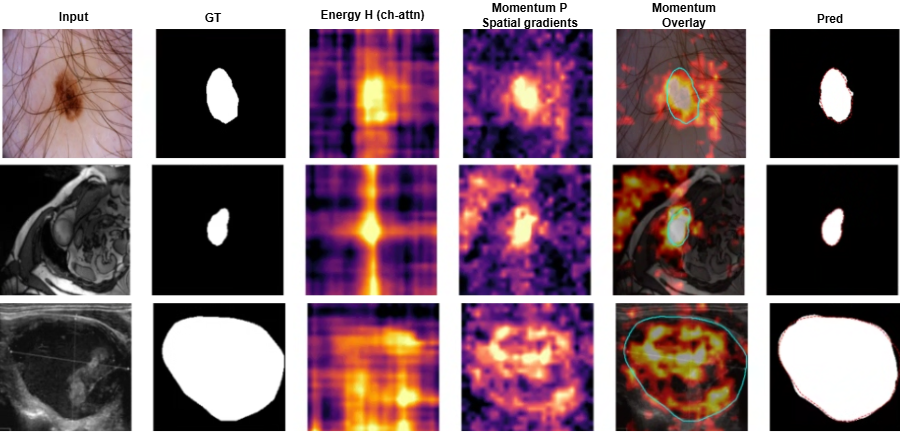}
\caption{Oscillator response across four datasets (rows: ACDC cardiac MRI, ISIC\,2018 dermoscopy, ISIC\,2017 dermoscopy, TN3K thyroid ultrasound). Left to right: input with ground truth overlay; energy-gated skip connections at decoder levels $d_3$, $d_2$, $d_1$ (note progressive spatial sharpening); momentum magnitude at $d_3$; momentum overlay. The skip gates highlight salient structures while suppressing background, and the momentum maps align with anatomical boundaries, all emerging from the Hamiltonian dynamics without boundary-specific supervision.}
\label{fig:physics_vis}
\end{figure*}

The energy $\mathcal{H}(i,j) = \tfrac{1}{2}(q^2 + p^2)$ integrates both position and momentum into a single scalar field. We measure the boundary-to-exterior energy ratio at 1.25 (Table~\ref{tab:diagnostics}), confirming that the energy map is elevated at boundaries and salient structures. The skip gate columns in Figure~\ref{fig:physics_vis} show how this energy map modulates encoder features at three decoder resolutions. At $d_3$ (coarsest, $28 \times 28$), the gate captures the overall spatial layout, which region is salient. At $d_2$ ($56 \times 56$), the gate sharpens to delineate individual structures. At $d_1$ ($112 \times 112$), the gate provides fine-grained boundary enhancement. This progressive refinement occurs because the energy map is computed at the bottleneck resolution and interpolated to each decoder level; the learned gain $\gamma_l$ at each level controls how aggressively the interpolated energy is used, with $\gamma_{d_3} = 1.18 > \gamma_{d_1} = 0.91$ indicating stronger gating at coarser levels where spatial selection is most critical.

\begin{figure*}[t]
\centering
\includegraphics[width=\textwidth]{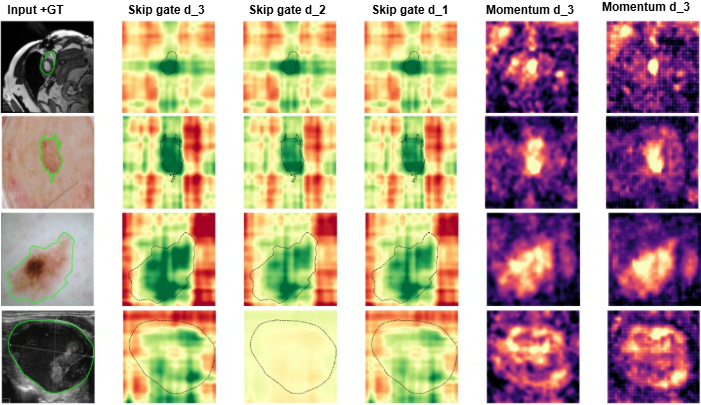}
\caption{Multiscale energy-gated skip connections across four datasets (rows: ACDC, ISIC\,2018, ISIC\,2017, TN3K). Columns show the gate activation $\sigma(\gamma_l(H_l - \bar{H}_l))$ at decoder levels $d_3$ (coarsest), $d_2$, and $d_1$ (finest). The progressive sharpening from diffuse spatial selection at $d_3$ to precise boundary delineation at $d_1$ demonstrates that the energy map, computed once at the bottleneck, provides complementary information at each decoder resolution. This coarse-to-fine pattern is consistent across all imaging modalities.}
\label{fig:multiscale_gates}
\end{figure*}

Figure~\ref{fig:multiscale_gates} provides a detailed view of this multiscale gating behavior across all four segmentation datasets. Each row corresponds to a different imaging modality, and the columns show the energy-gated skip connection response at decoder levels $d_3$, $d_2$, and $d_1$. The progression from coarse spatial selection to fine boundary delineation is clearly visible: at $d_3$, the gate activations are broad and diffuse, roughly delineating the region of interest; at $d_2$, the activations tighten around the structure; and at $d_1$, they sharpen into precise boundary-following contours. This behavior is consistent across all four modalities despite substantial differences in image characteristics, the cardiac chambers in ACDC, irregular lesion margins in ISIC, nodular boundaries in TN3K all exhibit the same coarse-to-fine gating pattern. The consistency confirms that the multiscale energy propagation captures a general property of the oscillator dynamics rather than modality-specific features.

\subsubsection{Do Phase-Space Statistics Transfer to Classification?}

For classification, the same physical quantities are globally pooled rather than spatially exploited. The pooled momentum $\bar{p} = \text{GAP}(|p|)$ captures the average spatial gradient activity across the image, a measure of texture complexity. The pooled energy $\bar{h} = \text{GAP}(\mathcal{H})$ captures the overall excitation level. To verify that these carry discriminative information beyond what standard feature pooling provides, we note the BloodMNIST result: HamCls achieves 98.85\% accuracy, exceeding MedMamba-S (98.4\%) by $+0.45\%$. BloodMNIST distinguishes eight types of blood cells that differ primarily in nuclear morphology and cytoplasmic texture, precisely the kind of fine-grained textural distinction that pooled momentum (which measures how spatially active the oscillator is) should capture. On DermaMNIST, where the seven classes include conditions with vastly different boundary characteristics (e.g., smooth melanocytic nevi vs.\ irregular dermatofibromas), the phase-space representation matches the best baseline (MedViT-S) on AUC while using substantially fewer parameters. On PathMNIST, the phase-space representation achieves superior discrimination of tissue types and pathological conditions, further validating that momentum and energy encode discriminative information for classification.

\section{Discussion}
\label{sec:discussion}

The results across ten benchmarks, five imaging modalities, and two distinct tasks point to a consistent picture: HamVision's strategy of constraining a network's bottleneck to follow Hamiltonian dynamics provides a broadly useful inductive bias for medical image analysis. It is worth reflecting on why this works, what it implies about the generality of the approach, and how it connects to concurrent developments in the field.

The oscillator bottleneck imposes a specific prior on feature processing: through frequency-selective, energy-conserving dynamics with input-dependent damping. This prior aligns with the requirements of both segmentation and classification, albeit for different reasons. For segmentation, the critical information lies at boundaries (captured by momentum), the relative importance of spatial locations varies (captured by energy), and features at different scales carry different information (captured by the filterbank). For classification, the discriminative information lies in the statistical distribution of these same quantities over the entire image: the global momentum profile captures texture complexity, the global energy profile captures saliency distribution, and the global feature profile captures content. The parameter efficiency, 7--9M parameters versus 16--105M for competitors, follows directly from this alignment: the physics provides three distinct functional outputs from a single dynamical process, eliminating dedicated modules that would otherwise require separate parameterization.

The generality across imaging modalities deserves emphasis. Segmentation spans dermoscopy (ISIC\,2018/2017), thyroid ultrasound (TN3K), and cardiac MRI (ACDC). Classification spans blood cell microscopy (BloodMNIST), colon pathology (PathMNIST), dermatoscopy (DermaMNIST), breast ultrasound (BreastMNIST), retinal fundus photography (RetinaMNIST), and abdominal CT imaging (OrganSMNIST). These datasets differ substantially in contrast, noise characteristics, resolution, and structural complexity. Yet the same oscillator bottleneck, with identical hyperparameters aside from the number of output classes or channels, achieves state-of-the-art or competitive results on all of them. This suggests that the oscillator captures something fundamental about visual analysis itself, the need to detect where features change, how active they are, and what frequency content they carry, rather than properties specific to any modality.

The connection to concurrent work in language modeling strengthens the theoretical grounding. Mamba-3~\cite{mamba3} independently proved that complex-valued state transitions are algebraically necessary for state-tracking tasks that real-valued SSMs cannot solve, and demonstrated significant language modeling improvements. The present work extends this insight by showing that the physical interpretation of the complex state, the decomposition into position, momentum, and energy, provides practical and interpretable structured representations for vision: these quantities serve as functionally distinct intermediate representations that improve segmentation through spatial gating and boundary injection, and classification through discriminative pooling. This is a contribution that Mamba-3, which treats complex transitions as an internal computational detail without considering their phase-space structure, does not address.

Several limitations merit discussion. The current architecture applies the oscillator only at the bottleneck resolution; extending it to earlier encoder stages or using multi-resolution oscillators is a natural direction. The row/column scanning strategy may miss diagonal boundary information. The exponential-Euler discretization we use is first-order; adopting the exponential-trapezoidal discretization introduced by Mamba-3~\cite{mamba3} could provide a more expressive recurrence with an implicit convolution on the state-input. Furthermore, while we have demonstrated HamVision on ten medical imaging datasets, evaluation on natural image benchmarks would test the generality of the oscillator hypothesis beyond the medical domain.

\section{Conclusion}
\label{sec:conclusion}

We have introduced HamVision, a unified framework for medical image analysis built on the damped harmonic oscillator as a structured inductive bias. The oscillator's phase-space decomposition yields position, momentum, and energy as functionally distinct intermediate representations that emerge from the dynamics rather than from dedicated learned modules. By combining a shared encoder and Hamiltonian bottleneck with task-specific heads, energy-gated skip connections with momentum injection for segmentation, and phase-space pooling for classification, we achieve state-of-the-art results on four segmentation benchmarks spanning three imaging modalities and state-of-the-art or competitive results on six classification benchmarks spanning five imaging modalities, all with 7--9M parameters.

The deeper insight of this work is that the choice of dynamical system imposes a prior on what the network can easily learn. Just as convolutions encode translation equivariance, the Hamiltonian oscillator encodes the prior that features evolve through frequency-selective, energy-conserving dynamics where spatial transitions produce structured boundary and saliency signals. The effectiveness of this prior across diverse modalities and tasks, combined with full interpretability of the intermediate representations, suggests that Hamiltonian dynamics constitute a principled and practical foundation for medical image analysis.

\section*{Data and Code Availability}

All datasets used in this study are publicly available. The segmentation benchmarks include ISIC\,2018, ISIC\,2017, TN3K, and ACDC. The classification benchmarks are drawn from MedMNIST v2. The source code, trained models, and configuration files for reproducing all experiments are available at \url{https://github.com/Minds-R-Lab/hamvision}.

\end{document}